# DLOL<sub>IS-A</sub>: Description Logic based Text Ontology Learning

Sourish Dasgupta
DA-IICT, India
sourish_dasgupta
@daiict.ac.in

Ankur Padia
DA-IICT, India
padiaankur
@gmail.com

Kushal Shah
DA-IICT, India
kushalshah40
@gmail.com

Prasenjit Majumder
DA-IICT, India
p_majumder
@daiict.ac.in

## ABSTRACT
Ontology Learning has been the subject of intensive study for the past decade. Researchers in this field have been motivated by the possibility of automatically building a knowledge base on top of text documents so as to support reasoning based knowledge extraction. While most works in this field have been primarily statistical (known as *light-weight Ontology Learning*) not much attempt has been made in axiomatic Ontology Learning (called *heavy-weight Ontology Learning*) from Natural Language text documents. Heavy-weight Ontology Learning supports more precise formal logic-based reasoning when compared to statistical ontology learning. In this paper we have proposed a sound Ontology Learning tool $DLOL_{IS-A}$ that maps English language IS-A sentences into their equivalent Description Logic (DL) expressions in order to automatically generate a consistent pair of ***T-box*** and ***A-box*** thereby forming both regular (definitional form) and generalized (axiomatic form) DL ontology. The current scope of the paper is strictly limited to IS-A sentences that exclude the possible structures of: (i) implicative IS-A sentences, and (ii) "*Wh*" IS-A questions. Other linguistic nuances that arise out of pragmatics and epistemic of IS-A sentences are beyond the scope of this present work. We have adopted Gold Standard based Ontology Learning evaluation on chosen IS-A rich Wikipedia documents.

## Categories and Subject Descriptors
I.2.4 [**Knowledge Representation Formalism and Methods**]: Representation (*procedural and rule-based*)

## 1. INTRODUCTION
The Ontology Learning is the automated generation of ontologies from documents that are primarily textual (i.e. Natural Language (NL) based). The primary intention is to build a knowledge base that formally reflects both assertive facts as well as general truth statements expressed in such text documents. A formal knowledge base allows reasoning based Information Retrieval (IR) and Extraction (IE), Question-Answering (QA), Machine Translation (MT), etc. that cannot be done on text [1]. Depending upon the degree of formalization Ontology Learning can be classified into two broad approaches: (i) Light-weight Ontology Learning and (ii) Heavy-weight Ontology Learning [1, 2]. Light-weight Ontology Learning leverages statistical NLP techniques and Machine Learning (ML) based clustering algorithms to generate conceptual hierarchies that may range from simple IS-A taxonomies to richer relational hierarchies. Linguistic analysis of texts is limited to concept labeling and sometimes relation extraction. On the other hand, heavy-weight Ontology Learning primarily depends upon linguistic analysis (essentially NL understanding) of texts and then transforming NL statements to predicate logic based expressions [3]. This results in the automatic generation of formal axiomatic ontologies as opposed to statistically generated informal ontologies.

The key arguments behind research efforts in heavy-weight Ontology Learning are: (i) formal ontologies support more accurate reasoning (assuming that the associated linguistic analysis is sound), (ii) formal ontology can be represented in Description Logics (DL) based languages like OWL[1] for ontology integration and mapping, and (iii) knowledge base management is easier for domain experts as well as ontology engineers. However, such an endeavor requires comprehensive linguistic analysis of languages in which texts are documented. Most languages have wide linguistic expressivity that makes NL understanding extremely difficult. Also linguistic pragmatics incorporates idiomatic and figurative expressions whose semantics cannot be understood literally using computational models such as Context-Free Grammar (CFG). In this paper we have proposed a linguistic analysis based heavy-weight Ontology Learning tool called $DLOL_{IS-A}$ (*Description Logic based Ontology Learning – ISA*) that leverages a comprehensive linguistic rule on a certain category of IS-A sentences where the core structure of an IS-A sentence is of the form: [Subject (S)] [IS – A] [Object (O)]. DLOL<sub>IS-A</sub> accepts simple, complex (i.e. clausal), and compound IS-A sentences, maps them to their corresponding linguistic structures and applies the linguistic rules associated with these structures in order to convert the sentences into their equivalent DL expressions. Although the current version of DLOL<sub>ISA</sub> cannot accept all kinds of complex IS-A sentences but we argue that the most common complex possibilities are covered. We understand that a complete ontology learning tool should also be able to cover non-ISA sentences. However, the thesis of this paper primarily focuses on the various non-trivial subtleties in IS-A sentences that one must be careful about while developing an axiomatic Ontology Learning tool. It has to be understood that the scope of the paper does not include: (i) Epistemic subtleties in IS-A sentences, (ii) idiomatic and figurative sentences – DLOL<sub>IS-A</sub> does complete literal interpretation, (iii) "*Wh*" IS-A questions – format: "*What is X?*", (iv) implicative IS-A sentences – format: "*If X is Y then P is Q.*" Our contribution in this paper can be summarized as follows:

1. Translation rules of IS-A sentences into T-Box definitions and A-Box assertions, thereby generating the regular (and corresponding generalized) T-Box.
2. Induction of A-Box assertions into respective T-Box definitions
3. Monotonic and Non-monotonic online revision of ontology
4. Evaluation of DLOL<sub>IS-A</sub> in terms of accuracy and efficiency

---
[1] http://www.w3.org/TR/owl-features/

The paper is organized into the following sections: (i) Related Work outlining some of the major contributions in light-weight and heavy-weight Ontology Learning, (ii) Problem Statement defining heavy-weight Ontology Learning formally, (iii) Approach in which IS-A sentences are formally characterized, T-Box and A-Box rule formulation proposed, complex/compound to simple sentence normalization rule formulation proposed, and ontology revision rules proposed, (iv) Methodology describing the $DLOL_{IS-A}$ architecture and algorithms related to text pre-processing, and recursive ontology learning, and (v) Evaluation in terms of accuracy.

## 2. RELATED WORK

There has been significant literature over the last decade on the problem of Ontology Learning. Most of these works can be categorized into two approaches as discussed earlier: (i) light-weight Ontology Learning, and (ii) heavy-weight Ontology Learning. We divide the discussion in this section into these two approaches.

### 2.1 Light-weight Ontology Learning

Light-weight ontology learning from text documents is the approach that is mostly taken by researchers in this field. Many of these works first goes through a rigorous statistical pre-processing of the given corpus. A very common pre-processing technique adapted by many is topic signature identification where a dataset is generated that is classified into topical senses [4, 5]. Out of the dataset concept formation is done using term extraction techniques (such as LSI [6], co-occurrence analysis, semantic lexicon referencing such as WordNet [7]. Some works have also used Formal Concept Analysis (FCA) for concept formation as well [8]. Finally, concept hierarchies are induced using unsupervised learning algorithms such as agglomerative hierarchical clustering [6, 9, 10] and variants of K-mean clustering [6]. In relation to unsupervised clustering different semantic distance measures (such as Cosine similarity, Jaccard similarity, Jensen-Shannon divergence, NGD [11]) have been proposed. There has also been works that have proposed non-taxonomic relation extraction and enrichment of learned ontologies [12, 13]. Some works used techniques that was very much domain specific and dependent on existing ontology templates [14]. A large portion of ontology learning techniques are semi-automated where user intervention is required for the final modeling and knowledge representation [15].

### 2.2 Heavy-weight Ontology Learning

Heavy-weight Ontology Learning, unfortunately, does not have much literature support if compared to light-weight ontology learning [16]. The main reason is that heavy-weight ontology learning is primarily dependent on NL understanding. This involves rigorous linguistic analysis and characterization of sentences in a particular natural language of interest. One of the early works in characterization of sentences can be attributed to the works in [7, 17]. In these works the primary focus is in identifying syntactic structural patterns and the inherent dependency of terms and relations. These patterns are then mapped to corresponding expressions in formal logic [18, 19]. Some works have used bootstrapping techniques to discover such patterns from corpus for enriching such axiomatic learning [20]. It has been observed in [6] that while the precision is quite high the recall is low since such pattern based characterizations were not adequate to cover all the variations in the underlying natural language. A few works can be found based on theoretical linguistic notions such as sub-categorization frame [21] to leverage linguistic association of a verb with a particular set of subject terms and object terms.

## 3. PROBLEM STATEMENT
### 3.1 Problem Overview

Heavy-weight Ontology Learning task involves three core tasks: (i) identifying the linguistic structure (i.e. lexico-syntactic pattern) of a given NL sentence, (ii) mapping the structure to the relevant linguistic rule, (iii) using the rule to define linguistic terms as predicate logic based equivalent expressions, and (iv) using satisfiability checker (i.e. reasoners) to construct the concept hierarchy (i.e. axiomatic ontology) over these predicate expressions.

In order to execute the first task we need to do some text pre-processing that may include: (i) Anaphora Resolution for resolving ambiguity in term (such as pronoun, proper noun, etc.) referencing, (ii) Part-Of-Speech (POS) Tagging for term extraction and linguistic pattern recognition, (iii) Parsing for normalizing clausal sentences into their corresponding simple form and disambiguation in semantic association of POS within syntactic structures. Anaphora resolution is itself a very difficult problem and there is significant scope of improvement for current anaphora resolver. However, the problem of Ontology Learning is independent from this problem since anaphora resolvers are used only as plug-ins by current Ontology Learning tools. POS Taggers are perhaps the most important plug-in for any axiomatic Ontology Learning tool. This is because such tools heavily depend on linguistic rules that can be applied correctly only if the linguistic structure of a NL sentence is identified accurately by a POS tagger. Incorrect POS tagging can mislead an axiomatic Ontology Learning tool to accept an unwanted (although correctly designed) rule for a given NL sentence, thereby generating an incorrect definition for the corresponding concepts. Sentence Parsers are yet another very important plug-in for Ontology Learning tools since NL sentences are mostly clausal in nature with POS dependencies within syntactic structures and other forms of linguistic ambiguity such as the *garden path sentences* like:"*John is watching the man with the telescope*". Here either of the nouns *John* and *Man* can be associated with the noun *telescope*. Although disambiguation in this case fairly depends on contextual information yet Parsers in general will give some default result. To sum it up, since our problem of Ontology Learning is independent of these three related problems it is just a matter of choice of selecting the best off-shelve tools by the Ontology Learning tool for improving its accuracy.

Inaccuracy in heavy-weight Ontology Learning itself is mainly innate to the second process of rule mapping. Linguistic rules are functions that transform identified linguistic structures into corresponding *equivalent* logic expressions. By equivalency we mean that the linguistic semantics of a given structure can be completely represented using an interpretation of the predicate logic that is used. If the linguistic semantics is incorrectly judged and/or there does not exist any interpretation in the chosen predicate logic that can be a model of the NL sentence then only the inaccuracy can be attributed to the ontology learning method. So the challenges of heavy-weight Ontology Learning are: (i) precisely identifying the linguistic semantic of NL sentences, and (ii) choosing a predicate logic variant that is expressive enough to have an interpretation for all possible variations of NL sentences. The scope of this current paper restricts such NL sentences to only IS-A forms.

## 3.2 Problem Formulation

Given a dynamic set of English IS-A sentences $S_{IS-A}$ model a transformation function $\tau_{IS-A}$ such that:

- $\tau_{IS-A}: S_{IS-A} \mapsto L_{IS-A}$
- $\forall s \in S_{IS-A} \; \exists l \in L_{IS-A} \ni l^{I_L} \equiv s^{I_{ENG}}$

where:

- $L_{IS-A}$ is a set of logic expressions of language $L$.
- $l$ is a logic expression of an English IS-A sentence $s$.
- $I_L$ is interpretation function for $L$.
- $I_{ENG}$ is linguistic interpretation function of $S_{IS-A}$.

# 4. APPROACH
## 4.1 General Outline

In our approach we choose the logic language to be Description Logics (DL). We argue that most IS-A sentences have expressive equivalency in the DL language: $\mathcal{AL}[\mathcal{U}][\mathcal{E}][\mathcal{C}][\mathcal{H}][\mathcal{O}]$ where:

- $\mathcal{AL}$: Attributive Language – supports atomic concept definition, concept intersection, full value restriction, role restriction, and atomic concept negation.
- $[\mathcal{U}]$: Union – supports concept union
- $[\mathcal{E}]$: Existential – supports full role restriction
- $[\mathcal{C}]$: Complement – supports concept negation
- $[\mathcal{H}]$: Role Hierarchy – supports inclusion axioms of roles
- $[\mathcal{O}]$: Nominal – supports concept creation of unrecognized Named Entity

Thus, the $L_{IS-A}$ set defined earlier (section 3.2) is fully definable with the language $\mathcal{AL}[\mathcal{U}][\mathcal{E}][\mathcal{C}][\mathcal{H}][\mathcal{O}]$. To model the transformation function $\tau_{IS-A}$ we first characterize the $S_{IS-A}$ set as well-formed linguistic structures. It has been observed that a significant number of structural possibilities within such characterization can be regarded as linguistically invalid (i.e. they do not generate any semantics). Out of the remaining valid structural possibilities we first give a model theoretic equivalent interpretation in terms of $\mathcal{AL}[\mathcal{U}][\mathcal{E}][\mathcal{C}][\mathcal{H}][\mathcal{O}]$ expressions. These expressions can be: (i) A-Box assertions and (ii) T-Box definitions or inclusion axioms.

Every A-Box statement must have a T-Box induction. T-box induction strictly follows the Open World Assumption (OWA). For an example, if a sentence is encountered such as: "*John is a player*" and if we can recognize the Named Entity *John* as a Person then although we cannot claim that "*Person is a player*" we cannot even disclaim it. Hence, the induction of the assertion "*John is a member of the class Player_Person*" will be "*Player_Person is a player*" and "*Player_Person is a Person*". However, we cannot induce that "*Some Person are not Player_Person*". T-box induction also follows the rule of *Most Specific Named Entity Recognition* (or *MSNER*) where if a Named Entity *X* is recognized as type *A* and type *B* and if type *B* is known to be subtype of type *A* then the *X* must be a asserted as a member of type *B* and the corresponding induction should be in terms of type *B*. For an example, if *John* is recognized as Male as well then instead of inducing "*Player_Person is a player*" we would have induced "*Player_Male_Person is a player*", "*Player_Male_Person is a Male_Person*", "*Male_Person is a Person*". The corresponding assertion would be: "*John is a member of the class Player_Male_Person*".

Both A-Box and T-box may require revision as new sentences are encountered. Revisions can be both monotonic and non-monotonic in nature. Monotonic revisions require only broadening the scope of a previous axiom or assertion. For an example, sentence 1: "*John is a doctor*"; sentence 2: "*John is a cardiologist*". These two sentences should not generate two separate assertions in the A-Box. Furthermore, the corresponding induced T-Box axiom must be single (assuming both the references of *John* is unambiguously identical). After sentence 2 has been encountered the revised version should be "*Cardiologist_Person is a cardiologist*" (from the earlier version of "*Doctor_Person is a doctor*") and a new induction of "*Cardiologist is a doctor*". The corresponding assertion revision will be "*John is a member of Cardiologist_Person*" (from the previous version of "*John is a member of Doctor_Person*"). Here we cannot, however, induce that "*Cardiologist_Person is a Doctor_Person*" because of OWA. On the other hand, non-monotonic revisions may have to done when an earlier axiom or assertion is no more valid. A famous example is the following sentence set: {sentence 1: "*Bird is a Flying Animal*", sentence 2: "*Penguin is a bird*", sentence 3: "*Penguin is not a Flying Animal*"}. Here non-monotonic revisions are required both in the T-Box and in the A-Box. Ontology revision will be detailed more formally in 4.3.

## 4.2 IS-A Sentence Transformation

There can be three general kinds of IS-A characterization: (i) simple (for simple IS-A sentences), (ii) complex (for clausal IS-A sentences), and (iii) compound (for conjunctive and disjunctive IS-A sentences). The fundamental characterization of any of these IS-A sentence types can be formalized as: **[S] [IS-A] [O]** where [S] denotes subject of the sentence, [O] denotes object of the sentence, and [IS-A] denotes the IS-A variations like: {"*is a/an*", "*is*", "*is kind of*", "*includes*", "*is class of*", "*is same as*" …} and other variations which can be included in the IS-A type like: {"*such as*", "*is like*" …}. Of course all past and future tense sub-variations and plural sub-variations of will also be necessarily included in the list. DLOL$_{IS-A}$ includes an IS-A list size of 26 different variations (excluding plural and tense sub-variations). The IS-A variation list is generated through an iterative bootstrapping process over the WordNet lexicon and then linguistically validated manually. It can be observed that IS-A variations can be of four categories: (i) subject hyponymy (ex: "*Cat is an animal.*"), (ii) subject hypernymy (ex: "*Animal includes cat.*"), (iii) subject membership (ex: "*John is a Human.*"), (iv) subject commonality (ex: "*Man is like machine.*"), and (v) subject quantification (ex: "*Students such as John and Joe are clever.*") While the first three categories are quite obvious the last two categories have linguistic subtleties that need to be addressed carefully. In the following sub-sections we characterize simple, complex, and compound sentences.

### 4.2.1 Simple IS-A Sentence Transformation

A simple IS-A sentence can be characterized as the sequence:
$[[Q_1]][[M]^*][S][IS-A][[Q_2]][[M]^*][O]$
where:

- [[ ]]: second square bracket indicates optional component
- []*: indicates multiple consecutive components of same type
- [$Q_1$]: subject quantifier – includes variations of the set: {a, an, the, some, all}
- [$Q_2$]: object quantifier – includes variations of the set: {the, some, all}
- [M]: subject/object modifier – value is restricted to the set: {NN, JJ, RB, VBG}
- [S]: subject – value is restricted to the set: {NN, NNP, JJ, RB, VBG}
- [O]: object - value is restricted to the set: {NN, NNP, JJ, RB, VBG}

- NT: set member notations follow the Penn Treebank).

The total number of structural possibilities for a simple sentence in the aforesaid characterization is 15,000 (without considering the variations of [IS-A] and [Q]). Out of these possibilities we observed that there are 4,601 linguistically valid possibilities. We now discuss each of the components in the following sub-sections.

### 4.2.1.1 Nuances of $[S][IS-A][O]$

Quantifier-free and modifier-free IS-A sentences are not trivial in all cases as it may seem. In this section we discuss the most important nuances and the rules to incorporate them in the transformation function $\tau_{IS-A}$ in the following.

**A. Subject-Object Dependency Problem:** A very important nuance of English language that significantly reduces the number of valid structures is the mutual dependency of the subject and the object. In IS-A sentences there are some kinds of pairs that can never occur together in terms of linguistic validity. Some examples are listed below:

- $[S = NNP][IS - A][O = RB]$ – A named entity cannot be a member of an adverb.
- $[S = RB][IS - A][O = NNP]$ – An adverb cannot be a class of or same as a named entity.
- $[S = NNP][IS - A][O = VBG]$ – A named entity cannot be a member of an activity.
- $[S = VBG][IS - A][O = NNP]$ – An activity cannot be a class of a named entity.
- $[S = JJ][IS - A][O = RB]$ – An adjective cannot be a sub class or class of an adverb.
- $[S = JJ][IS - A][O = VBG]$ – An adjective cannot be a sub class or class of an activity.

We hereby see that out of a total of 25 core structural possibilities 6 are eliminated. This leads to a total elimination of 3600 structures (out of 15,000 structures).

**B. Membership versus Inclusion Problem**: Almost all kinds of IS-A sentences are either inclusion statements (ex: "*Man is an Animal*") or membership statements (ex: "*John is a Man*"). Membership can happen only when the S is an NNP (Proper Noun) and the O is a NN (Noun). This is because of: (i) there cannot be a named instance of a concept which is either a VBG (Gerund) or an RB (adverb) or a JJ (adjective), (ii) the O cannot have an instance value including NNP if the IS-A variation is not inverse to "is a". Examples of inverse variations are "includes", "is class of", etc., and (iii) if both S and O is NNP then the IS-A variation essentially means "is same as". Example sentence is "*John is Joe*". Under such circumstances the corresponding A-Box assertion and its induced T-box are:

- **A-Box Rule:** If $[S = NNP][IS - A][O]$ Then $O(S)$; $(WordNet.getMSP(S))(S)$;
- **T-Box Induction Rule:** If $[S = NNP]$ Then $['O' + 'WordNet.getMSP(S)'] \sqsubseteq O$; $['O' + WordNet.getMSP(S) \sqsubseteq [WordNet.getMSP(S)]$;

where *getMSP* gives the Most Specific Parent of S from WordNet. If the MSP is not found then:

- **T-Box induction rule**: $['O'+'\{S\}'] \sqsubseteq O; ['O' + '\{S\}'] \sqsubseteq \{S\}$

where *{S}* is the nominal of the subject named entity. All other cases are candidates of inclusion axiom in the T-box of the nature:
- **T-Box Rule**:
  If $[IS - A] \neq inverseOf("is\ a")$ Then $S \sqsubseteq O$; Else $O \sqsubseteq S$

It has to be understood that some of these candidates having "same as" variations (7 enlisted) are essentially equivalent axioms. For such cases :
- **T-Box rule:** $S \equiv O$

**C. Object Reification Problem:** In many valid structures it is difficult to directly transform the sentences into their equivalent DL form. This happens when [O] assumes the form: $[O \in \{JJ, RB\}]$ while [S] assumes the form: $[S \in \{NN, NNP, VBG\}]$. For an example, the sentence: "*John is beautiful*" does not mean that "John" is a member of the concept "Beautiful". To solve this problem we introduce two primitive concepts: (i) **Attribute** (for JJ and RB) and (ii) **Activity** (for VBG). All types of JJ and RB are sub concepts of the primitive concept **Attribute**. Similarly all types of VBG are sub concepts of the primitive concept **Activity**. Each of these two primitives has two associated primitive roles for which they act as fillers: (i) **hasState** (for **Attribute**) and (ii) **does** (for **Activity**). The second primitive role will not be required for IS-A sentences though. We then apply a general reification rule for structure $[S = NN/NNP][IS - A][O = JJ/RB]$ as follows:

- **T-Box Rule**: $S \sqsubseteq ['O' + 'Thing']$; $['O' + 'Thing'] := \forall hasState.O; O \sqsubseteq Attribute$;

Similarly, for the structure $[S = VBG][IS - A][O = JJ/RB]$ the following rule is applied:
- **T-Box Rule:** $S \sqsubseteq ['O' + 'Activity']$; $['O' + 'Activity'] \sqsubseteq Activity$;

**D. "LIKE" Problem**: The IS-A variation "like" and its variants (10 enlisted) pose a very interesting problem since its linguistic semantics is not exactly same as either "is a" or "same as". For an example, a sentence such as "*Apple is like Orange*" does not imply that the concept Apple is either equivalent to or sub concept of the concept Orange. What it means is that the subject concept Apple and the object concept Orange share some common characteristics and can be grouped under one concept representing the commonality. The corresponding rule is:

- **T-Box Rule**: $S \sqsubseteq ['O' + 'Like']; O \sqsubseteq ['O' + 'Like']$;

**E. "SUCH AS" Problem**: The IS-A variation "such as" and its positional variations [7] also poses two very interesting semantic deviations. In some sense "such as" behaves like a modifier to the subject of a IS-A sentence. It is in respect to this behavior that "such as" can come in two flavors: (i) conjunctive and (ii) disjunctive. An example of conjunctive form is the sentence: "*Students, such as John and Joe, are intelligent*". Its disjunctive variation will be the sentence: "*Students, such as John or Joe, are intelligent*". In both the cases we are not talking about the entire concept Student being sub concept of the concept IntelligentThing. But in the conjunctive form both John and Joe should belong to the restricted concept of IntelligentStudent while in the disjunctive form either John or Joe have to belong to the restricted concept IntelligentStudent. Hence, the distinction between the two cases is given in the following rule:

- **T-Box Rule for** $[S][such\ as][\{\wedge O_{1i}\}][IS - A][O_2]$: $[\{'O'_{1i}\} + 'S'] \sqsubseteq O_2; [\{'O'_{1i}\} + 'S'] \sqsubseteq ['O'_2 + 'S']$

In the given example, the corresponding axiom will be:
$JohnJoeIntelligentStudent \sqsubseteq IntelligentThing$;
$JohnJoeIntelligentStudent \sqsubseteq IntelligentStudent$;
$IntelligentStudent \sqsubseteq Student$

- **A-Box Rule**: $JohnJoeIntelligentStudent(John)$; $JohnJoeIntelligentStudent(Joe)$;
- **T-Box Rule for** $[S][such\ as][\{\vee O_{1i}\}][IS - A][O_2]$: $[['O'_{11} +'O'_2 +'\ S'] \sqcup ... \sqcup ['O'_{n1} +'O'_2 +'\ S']] \sqsubseteq O_2$; $['O_{1i}' +'S'] \sqsubseteq ['O'_2 + 'S']$

In the given example, the corresponding axiom will be:
$(JohnIntelligentStudent \sqcup JoeIntelligentStudent) \sqsubseteq IntelligentThing$;
$JohnIntelligentStudent \sqsubseteq IntelligentStudent$;
$JoeIntelligentStudent \sqsubseteq IntelligentStudent$;
$IntelligentStudent \sqsubseteq Student$
- **A-Box Rule:** $JohnIntelligentStudent(John)$; $JoeIntelligentStudent(Joe)$;

**F. Past Tense Ontological Ambiguity**: The past tense variation of "is a" is poses a difficult ontological ambiguity. To illustrate this we take the example of the sentence: "*Mammoths were huge*". In this case, the subject concept (i.e. Mammoth) does not exist anymore. However, the ontological validity of the subject will never change. In other words mammals will always remain huge with respect to any given time point. However, this is not true for all sentences with equivalent characterization. For an example: "*Human was uncivilized*". This actually means that the subject concept Human is no more uncivilized. In other words, the ontological validity is not applicable. However, this can only be said if we know that the subject concept has a different ontological validity at present. This is obviously a very difficult problem by itself. DLOL$_{IS-A}$ do not attempt to solve this problem. Instead, it takes a "play safe" attitude to guarantee at least partial ontological validity of such sentences. In this approach the subject can be either what it was before or otherwise. Hence, Mammoth can be a HugeThing or a non HugeThing. However, this would mean that Mammoth is a kind of anything which is a tautology and hence, adds no new information to the knowledge base. Thus, Mammoth must have a primitive role called **Past Pointer Role** (**PPR**) that states that it is a subset of something that is either a HugeThing or has the **PPR** relation to HugeThing. The corresponding rule is as follows:

- **T-Box Rule**: $S \sqsubseteq (O \sqcup \forall PPR.O); (O \sqsubseteq \neg (\forall PPR.O)) \equiv \bot$
If $[S = NNP]$ Then
- **A-Box Rule:** $(['PPR'+'O' + WordNet.getMSP(S)])(S); [WordNet.getMSP(S)](S)$;
- **T-Box Induction Rule**:
$['PPR'+'O' + WordNet.getMSP(S) \sqsubseteq (O \sqcup \forall PPR.O);$
$(O \sqsubseteq \neg (\forall PPR.O)) \equiv \bot;$
$['PPR' + 'O' + WordNet.getMSP(S)] \sqsubseteq [WordNet.getMSP(S)]$;

### 4.2.1.2 Nuances of Quantifiers

[Q] acts as a determiner (DT) in a sentence. Hence, the $[Q_1][S]$ sub-form contextualizes the subject. Similarly, the $[Q_2][O]$ sub-form contextualizes the object. Hereby we discuss the most important nuances and the rules to incorporate them in the transformation function $\tau_{IS-A}$ in the following.

**A. Subject Contextualization**: An interesting linguistic nuance that can be observed about the structure: $[Q_1 = null][S][IS - A][Q_2 \in \{the, some\}][O]$ is that since O is contextualized the S also gets contextualized as a consequence. For an example, in the sentence "*Animal is the victim.*" we cannot say for sure that the sentence is true for the entire animal class. Most likely, a particular animal that has been discussed in some previous context is being referred here. If co-reference resolution successfully identifies the animal then the sentence will be treated by DLOL$_{IS-A}$ as equivalent to the sentence "*The animal is the victim.*"; otherwise it will be treated as "*Animal is a victim*".

**B. Epistemic Ambiguity**: Another linguistic nuance of [Q] is innate within the form: $[Q_1 \in \{a, an\}][S][is][O]$. Sentences in this form can be very ambiguous in terms of the subject's epistemic scope. For an example, a sentence such as "*A planet is round*" speaks of a general truth about all planets. However, another sentence such as "*A boy is hungry*" certainly does not speak of the entire boy class being hungry. Since both these sentences have same characterization DLOL$_{IS-A}$ takes a "pessimistic" attitude to guarantee at least partial epistemic validity of such sentences and treats the first sentence to be the same as the sentence "*The planet is round*" with a indefinite sense of determiner "the". This approach essentially leads to the induced (sometimes partial) truth that "*Some planets are round*". The corresponding rule is as follows:

- **T-Box Rule**: $['O' + 'S'] \sqsubseteq O; ['O' + 'S'] \sqsubseteq S$;
If $[S = NNP]$ Then
- **A-Box Rule:** $(['O' + WordNet.getMSP(S)])(S); [WordNet.getMSP(S)](S)$;
- **T-Box Induction Rule**:
$['O' + WordNet.getMSP(S)] \sqsubseteq O;$
$['O' + WordNet.getMSP(S)] \sqsubseteq S;$
$['O' + WordNet.getMSP(S)] \sqsubseteq [WordNet.getMSP(S)]$;

**C. "A" Variation Problem**: The third linguistic nuance of [Q] can be attributed to the wide range of lexical variations (or alternatives) to express same or similar quantification of subject or object. We have enlisted a variation list of size 703 which again has been generated from an iterative bootstrapping process over WordNet. It is to be noted that in the context of quantifiers [Q]* is considered to have a reflexive semantics in the sense: [Q]* = [Q] where Q is the last quantifier in the sequence. For an example, "*Many many mammals are omnivorous*" is considered semantically equivalent to "*Many mammals are omnivorous.*" Also, it is obvious that all variations of the quantifier "all" (11 enlisted) in IS-A sentences can be ignored. In other words, sentence such as "*Every cat is a mammal*" is semantically same as the sentence "*Cat is a mammal*". Furthermore, not all variations of the quantifier "a" are trivially same as that of the treatments of indefinite DTs "a/an" as discussed before. More precisely there are 4 enlisted variations (out of 6) that require special treatments – "only", "any one of", "one of", and "only one of".

**C.1 "ONLY" Problem**: The quantifier "only" can be found in the following three structures:

$[S][IS - A] [ONLY][O];$
$[ONLY][S][IS - A][O];$
$[ONLY][S][IS - A][ONLY][O]$

In the first structure the subject S has to be either a member or a sub-concept of the object concept O and nothing else. Hence, the following rule has to be made:

- **T-Box rule**: $S \sqsubseteq O; (S \sqcap \neg O) \equiv \bot;$
If $[S = NNP]$ Then
- **A-Box Rule**:
$(['ONLY' + O' + WordNet.getMSP(S)])(S); [WordNet.getMSP(S)](S)$;
- **T-Box Induction Rule**:
$['ONLY'+'O' + WordNet.getMSP(S)] \sqsubseteq O;$
$['ONLY'+'O' + WordNet.getMSP(S)] \sqsubseteq [WordNet.getMSP(S)];$
$(['ONLY' +' O' + WordNet.getMSP(S)] \sqcap \neg O) \equiv \bot;$

In the second structure no other concept other than S is a sub-concept of the object concept O. Hence, the following rule has to be made:

- **T-Box rule**: $S \sqsubseteq O; (\neg S \sqcap \neg O) \equiv \bot;$
If $[S = NNP]$ Then

- **A-Box Rule**:
  $(['O' +' ONLY' + WordNet.getMSP(S)])(S);$
  $[WordNet.getMSP(S)](S);$
- **T-Box Induction Rule**:
  $['O'+' ONLY' + WordNet.getMSP(S)] \sqsubseteq O;$
  $['O'+' ONLY' + WordNet.getMSP(S)] \sqsubseteq$
  $[WordNet.getMSP(S)];$
  $(\neg['O' +' ONLY' + WordNet.getMSP(S)] \sqcap \neg O) \equiv \bot;$

In the third structure both the above epistemic has to hold. Hence, the following rule has to be made:

Hence, the following rule has to be made:

- **T-Box rule**: $S \sqsubseteq O; (S \sqcap \neg O) \equiv \bot; (\neg S \sqcap \neg O) \equiv \bot$

If $[S = NNP]$ Then
- **A-Box Rule**:
  $(['ONLY'+'O'+'ONLY' + WordNet.getMSP(S)])(S);$
  $[WordNet.getMSP(S)](S);$
- **T-Box Induction Rule**:
  $['ONLY'+'O'+'ONLY' + WordNet.getMSP(S)] \sqsubseteq O;$
  $['ONLY'+'O'+'ONLY' + WordNet.getMSP(S)]$
  $\sqsubseteq [WordNet.getMSP(S)];$
  $(['ONLY' +' O' +' ONLY' +$
  $WordNet.getMSP(S)] \sqcap \neg O) \equiv \bot;$
  $(\neg['ONLY' +' O' +' ONLY' +$
  $WordNet.getMSP(S)] \sqcap \neg O) \equiv \bot;$

**C.2 "ANY ONE OF/ ONE OF" Problem**: The quantifiers "any of", "any one of", "one of", "only one of" can be found in the following three structures:

$[S][IS-A] [X][\{O\}];$
$[X][\{S\}][IS-A][\{O\}];$
$[X][\{S\}][IS-A][X][O]$
where $[X] \in \{[ANY\ ONE\ OF], [ONE\ OF], \}$

In the first structure the subject S has to be either a member or a sub-concept of the union of the set of object concepts {O}. However S cannot be a sub-concept of more than one of O. Hence, the following rule has to be made:

- **T-Box Rule**: $S \sqsubseteq \sqcup(O_i); \sqcap(S \sqcap \neg(O_i \sqcap O_j) \equiv \bot);$

If $[S = NNP]$ Then
- **A-Box Rule**:
  $(['ONEOF'+concat(\{'O_i'\}) +$
  $WordNet.getMSP(S)])(S);$
  $[WordNet.getMSP(S)](S);$
- **T-Box Induction Rule**:
  $\sqcup([(O_i) + WordNet.getMSP(S)]) \equiv ['ONEOF' +$
  $concat(\{'O_i'\} + WordNet.getMSP(S)];$
  $['O_i' + WordNet.getMSP(S)] \sqsubseteq$
  $WordNet.getMSP(S);$
  $['ONEOF' + concat(\{'O_i'\} + WordNet.getMSP(S)] \sqsubseteq$
  $WordNet.getMSP(S);$
  $['ONEOF' + concat(\{'O_i'\} + WordNet.getMSP(S)] \sqsubseteq$
  $\sqcup(O_i);$
  $\sqcap(['ONEOF' + concat(\{'O_i'\} + WordNet.getMSP(S)]$
  $\sqcap \neg(\sqcap(O_i, O_j) \equiv \bot);$

In the second structure only one of the member subjects in the set {S} can be a subset of O. Hence, the following rule has to be made:
- **T-Box Rule**: $\sqcup(S_i) \sqsubseteq O; \sqcap((S_i \sqcap S_j) \sqcap \neg O) \equiv \bot);$
If $[S = NNP]$ Then
- **A-Box Rule**:

$(['O' +' ONEOF'$
$+ leastCommonMSP(WordNet.getMSP(S_i))])(S_i);$
$[WordNet.getMSP(S_i)](S_i);$
- **T-Box Induction Rule**:
  $\sqcup(['O' + WordNet.getMSP(S_i)])$
  $\equiv ['O' +' ONEOF'$
  $+ leastCommonMSP(WordNet.getMSP(S_i))];$
  $['O' + WordNet.getMSP(S_i)] \sqsubseteq [WordNet.getMSP(S_i)];$
  $['O' +' ONEOF'$
  $+ leastCommonMSP(WordNet.getMSP(S_i))] \sqsubseteq O;$
  $\sqcap(['O' + WordNet.getMSP(S_i)] \sqcap ['O'$
  $+ WordNet.getMSP(S_j)]) \sqcap \neg O) \equiv \bot);$

In the third structure both the above epistemic has to hold. Hence, the following rule has to be made:

- **T-Box Rule**: $\sqcup(S_i) \sqsubseteq \sqcup(O_j); \sqcap(\sqcup(S_i) \sqcap \neg(O_i \sqcap O_j) \equiv \bot$
  $); \sqcap((S_i \sqcap S_j) \sqcap \neg O_i) \equiv \bot);$
If $[S = NNP]$ Then
- **A-Box Rule**:
  $(['ONEOF'+concat(\{'O_i'\}) +' ONEOF'$
  $+ leastCommonMSP(WordNet.getMSP(S_j))])(S_j);$
  $[WordNet.getMSP(S_i)](S_i);$
- **T-Box Induction Rule**:
  $\sqcup(['ONEOF'+concat(\{'O_i'\}) + WordNet.getMSP(S_j)])$
  $\equiv ['ONEOF'+concat(\{'O_i'\}) +' ONEOF'$
  $+ leastCommonMSP(WordNet.getMSP(S_j))];$
  $['ONEOF'+concat(\{'O_i'\}) + WordNet.getMSP(S_j)]$
  $\sqsubseteq ['ONEOF'+concat(\{'O_i'\})];$
  $['ONEOF'+concat(\{'O_i'\})] \sqsubseteq \sqcup(O_i);$
  $['O' + WordNet.getMSP(S_i)] \sqsubseteq [WordNet.getMSP(S_i)];$
  $['ONEOF'+concat(\{'O_i'\}) +' ONEOF'$
  $+ leastCommonMSP(WordNet.getMSP(S_i))]$
  $\sqsubseteq ['ONEOF'+concat(\{'O_i'\})];$
  $\sqcap(\sqcup(['ONEOF'+concat(\{'O_i'\}) + WordNet.getMSP(S_j)])$
  $\sqcap \neg(O_i, O_j) \equiv \bot);$
  $\sqcap(['O' + WordNet.getMSP(S_i)] \sqcap ['O'$
  $+ WordNet.getMSP(S_j)]) \sqcap \neg O_i) \equiv \bot);$

**C.3 "ONLY ONE OF" Problem**: The quantifier "only one of" is an extended case of the previous problem where it also inherits the nuances of the "ONLY" problem. Thus, for each of the three cases of the "ONE OF" rules we need to add the following rules respectively:

$[S][IS-A] [ONLY\ ONE\ OF][\{O\}] - S \sqcap \neg(\sqcup(O_i)) \equiv \bot$
$[ONLY\ ONE\ OF][\{S\}][IS-A][O] - \neg(\sqcup(S_i)) \sqcap \neg O \equiv \bot$
$[ONLY][S][IS-A][ONLY][O] - \neg(\sqcup(S_i)) \sqcap \neg(\sqcup(O_i)) \equiv \bot$

**D. Degree of Quantification Problem:** The fourth nuance of [Q] is due to the degree of quantification that [Q] variations pose. Grossly there can be four degrees: (i) high intensity (ex: "a lot of", "many", "most", etc), (ii) low intensity (ex: "not much", "hardly any", "few", etc), (iii) neutral intensity (ex: "some"), and (iv) numeral (ex" "just about five", "seven of", "approximately five", etc). Even though IS-A sentences with same subject and object bear different semantics when a quantifier of one type of intensity is replaced by another the DL interpretation function is limited in terms of capturing such semantic differences. For an example, the sentence "*Many cats are yellow*" is different than the sentence "*A few cats are yellow*". In the first case there is a *YellowCat_1* concept whose cardinality is very high. In the second case there is another *YellowCat_2* concept whose cardinality is low. However, both the concepts are DL equivalent since both must have the same set of characteristics (i.e. color being yellow) and concept cardinality is not expressible. This may be seen as a shortcoming of the underlying DL based approach. However, we stress that cardinality is not very important in this case since cardinality matters only in the A-Box where queries on cardinality can be given to the learned ontology. In this case since we cannot assert anything in the A-Box (no concrete instance is known) we do not have to worry about cardinality. Also, the notion of high, low, and neutral intensity is very subjective and depends on the domain as well as the author of a sentence. Hence, we cannot say that if $x$ number of instances are inserted consistently in any subject concept S then we have achieved a high/low/neutral intensity and hence, should consider the insertion of $(x+1)$th instance as inconsistent w.r.t the T-Box. It is because of these two reasons that DLOL$_{IS-A}$ ignores the quantification degree of IS-A sentences.

### 4.2.1.3 Nuances of Modifier

Normally, if a modifier in simple IS-A sentence is a JJ or an NN then it modifies either an NN or an NNP. For an example, in the sentence: "*Wild cat is a mammal*" the JJ "Wild" modifies the subject concept "Cat" which is a NN. In such general cases it is evident that the concept "WildCat" is a sub concept of the concept "Cat" and also is a sub concept of the concept "Mammal". However, modifiers can add to the nuances that we have seen so far in simple IS-A sentences. Here we discuss some of such nuances:

**A. Outward Unfolding Problem**: For sequence of modifiers such JJ and NN we can easily adopt an *outward unfolding* rule. The rule states that if, as an example, we have a structure such as $[M_1][M_2][M_3][S][IS-A][O]$ then it is interpreted as: $\left[M_1\left[M_2[[M_3][S]]\right]\right][IS-A][O]$ where $M_2$ modifies the concept "$M_3S$" and $M_1$ modifies the concept "$M_2M_3S$". The following rule is as follows:

- **T-Box Rule** $([M_n]\ldots[M_2][M_1][S][IS-A][O])$:
  $[concat(\{'M_i'\})+'O'+'S'] \sqsubseteq S$;
  $[concat(\{'M_i'\})+'O'+'S'] \sqsubseteq [concat(\{'M_{i-1}'\})+'O'+'S'] \sqsubseteq \cdots \sqsubseteq [concat(\{'M_1'\})+'O'+'S'] \sqsubseteq ['O'+'S'] \sqsubseteq O$;

**B. The Gerund Problem**: When gerunds (VBG) act as modifiers the rule of outward unfolding cannot be applied as it is. This is because in structures such as $[M=VBG][S=NN][IS-A][O]$ the modification can unfold either outward or inward. For an example, in the sentence "*Playing soccer is healthy*" the VBG modifier "Playing" is not modifying soccer since the concept PlayingSoccer cannot be sub concept of the concept Soccer. Instead, it is the concept Soccer that is modifying the concept Playing where SoccerPlaying is a kind of Playing. However, in another example sentence "*Running water is beautiful*" the outward unfolding rule holds true. The underlying ambiguity is very difficult to clarify since, unless mentioned, there is no way to understand whether the NN is an actor of the VBG or is acted upon (where the action is the VBG). In the previous example soccer is not playing but is played. While in the second example, water runs. We treat such sentences as follows:

- **T-Box Rule**: $['O'+'M'+'S'] \sqsubseteq (['O'+'Thing'])$;
  $['O'+'M'+'S'] \sqsubseteq (M \sqcup S)$;
  $['O'+'Thing'] \coloneqq \forall hasState.Attribute$;

**C. Problem of Qualified IS-A**: Sometimes certain types of RB modifies the "is a" component of a structure of the form: $[M=RB][S][IS-A][O]$ or $[S][IS-A][M=RB][O]$. Such kinds of modification creates an ontological ambiguity\that cannot be captured in classical DL. For an example, in the sentence "*Eventually sun is a black hole*" the subject concept Sun is not a black hole as of now but is in the process of becoming in the future. In such cases we introduce a primitive role called **Future Pointer Role** (**FPR**) that states that Sun is a subset of something that is eventually black hole (called EventuallyBlackHole) and has the **FPR** relation to BlackHole. The corresponding rule is as follows:

- **T-Box Rule**: $[S \sqsubseteq (['M'+'O'] \sqcup \forall FPR.O);$
  $(['M'+'O'] \sqsubseteq \neg(\forall FPR.O)) \equiv \bot$;

### 4.2.2 Complex IS-A Sentence Transformation

Complex IS-A sentences are clausal sentences where every clause is in the IS-A form. An example sentence is: "*A Predator is an animal that is animal-eater*". In this sentence the clause: "*an animal that is animal-eater*" is of type IS-A. We can characterize most forms of a complex IS-A sentence as:

$[[X]][S][[That]][IS-A][[X]][O_1][[That]][IS-A][[X]][O_2]$;

Where:

- $[[X]]: [[Q]][[M]^*]$
- [**That**]: signifies clausal token and all its variations (such as: 'which', 'who', 'where', etc.).

Some clausal tokens such as "yet" and "but" do not come in this characterization. Such sentences have S as the subject of both the clauses and the rule is same as that of form 2 given next. Depending upon the position of the clausal token 'That' we can have 4 T-box rules for the following cases:

$[S][IS-A][O_1][IS-A][O_2]$: This is an ambiguous form since it is difficult to determine the subject of the clauses. Also, the linguistic validity of the sentence can be debated as well. Under such circumstances we will assume that the linguistic semantics of the form is the same as the next form.

$[S][IS-A][O_1][That][IS-A][O_2]$: The position of the clausal token in this form makes $O_1$ the subject concept of the second clause while S remains the subject concept of the first clause. Example sentence is: "*Cat is a feline which is an animal.*" The corresponding rule is:

- **T-Box Rule** (**Modifier/Quantifier Free**): $O_1 \sqsubseteq O_2; S \sqsubseteq O_1$;
- **T-Box Rule** (**Object 1 Modifier**): $['M'+'O_1'] \sqsubseteq O_2$;
  $S \sqsubseteq ['M'+'O_1']; ['M'+'O_1'] \sqsubseteq M; ['M'+'O_1'] \sqsubseteq O_1$;
- **T-Box Rule** (**Object 1 Quantifier**): $['O_2'+'O_1'] \sqsubseteq O_2$;
  $['O_2'+'O_1'] \sqsubseteq O_1; S \sqsubseteq ['O_2'+'O_1']$;

$[S][That][IS-A][O_1][IS-A][O_2]$: The position of the clausal token in this form makes S the subject concept of both the clauses. Example sentence is: "*A cat that is Persian is long-haired.*" The corresponding rule is:

- **T-Box Rule** (**Modifier/Quantifier Free**): $S \sqsubseteq O_1; S \sqsubseteq O_2$;

Rules for first object modifier/quantifier follow the same principle as the previous form.

$[S][That][IS-A][O_1][That][IS-A][O_2]$: This form is again linguistically invalid since both the clause subjects (S and $O_1$) are unsaturated.

### 4.2.3 Compound IS-A Sentence Transformation

Compound IS-A sentences are IS-A sentences in the conjunctive or disjunctive list of the subject or the object or both. The basic generic format of such sentences can be characterized as:

$[[X]][\wedge S_i / \vee S_i][IS-A][[X]][\wedge O_j / \vee O_j]$

Where:

- $[X]: [[Q]][[M]^*]$
- $\vee / \wedge$ : Disjunctive/Conjunctive operator representing 'or' and their variations (ex: "either .. or ..", "as well as", etc.)
- $\wedge Y \in \{S, O\}$: Conjunctive list of subject/object
- $\vee Y \in \{S, O\}$: Disjunctive list of subject/object

More complex format can be recursively characterized as:

$[[X]][\wedge S_{1..k} / \vee S_{1..k}][[\vee / \wedge]][[\wedge S_{k..n} / \vee S_{k..n}]][IS-A][[X]][\wedge O_{1..l} / \vee O_{1..l}][\vee / \wedge][[\wedge O_{l..m} / \vee O_{l..m}]]$

An example disjunctive sentence of such format is: "*Cat, dog, and bull are either herbivorous or carnivorous.*" The corresponding rules are as follows:

$[\wedge S_{1..k}][IS-A][O]$: This sentence can be split into k simple sentences of the form $[S_i][IS-A][O]$ and hence follows the corresponding transformation rule.

$[\vee S_{1..k}][IS-A][O]$: The sentence semantics is same as the form $[ANYOF][S_i][IS-A][O]$ and follows the T-Box rule: $\sqcup S_i \sqsubseteq O$.

$[S][IS-A][\wedge O_{1..l}]$: This sentence can be split into $l$ simple sentences of the form $[S][IS-A][O_j]$ and hence follows the corresponding transformation rule.

$[S][IS-A][\vee O_{1..l}]$: The sentence semantics is same as the form $[S][IS-A][ANYOF][O_j]$ and follows the T-Box rule: $S \sqsubseteq \sqcup O_j$.

Rules for all other variations can be derived from these rules.

### 4.2.4 IS-NOT Sentence Transformation

Transforming sentences having negation is yet another nuance that NL poses. This is because of two reasons: (i) DL does not have interpretation for negation of relations (ex: "*Man does not like raw meat*") and (ii) simple negation of object concept does not fully capture the ontological as well as epistemic semantics of such sentence (ex: "*Man is not cat*" does not only imply that the subject concept Man is a sub concept of everything other than cat). The first problem is currently outside the scope of discussion of this paper since they do not appear in IS-A sentences. The second sentence can be characterized as: $[[NO]][[X]][S][IS-A][[NOT]][O]$ where either the 'NO' or the 'NOT' element has to be present (but not both). To capture the full epistemic we introduce yet another primitive role **hasProperty** such that **hasProperty ⊑ hasState; hasProperty ⊑ does**. We also introduce a placeholder primitive concept $\overline{S}$ (can be interpreted as the intrinsic characteristic of the subject S) that is a proper sub concept of ⊤. The corresponding rule is as follows:

**T-Box Rule:** $S \sqsubseteq \neg O; S \coloneqq \exists hasProperty.\overline{S}; B \sqsubset \top; \overline{S} \not\equiv \bot;$

## 4.3 Online Ontology Revision

The T-Box and A-box that is generated using the transformation function $\tau_{IS-A}$ may not be consistent as new contradicting sentences are encountered by DLOL$_{IS-A}$ over time. This requires DLOL$_{IS-A}$ to revise existing T-Box and A-Box non-monotonically (as has been discussed in section 4.1). Also, new sentences can add to old information about a particular concept requiring monotonic modification of T-Box definitions and/or A-Box assertions. In this section we propose online revision rule that DLOL$_{IS-A}$ uses. Such online rules are applied dynamically when a particular transformation rule is chosen. The rules are as follows:

I. **Non-monotonic Revision Rules**

A. **Rule 1**: If $Rule_{new}: C \sqsubseteq D$ and $(C \sqsubseteq \neg D) \in$ TBox Then

   **T-Box Revision:**

   $['D' + 'C'] \sqsubseteq D; ['NOT' +' D' + 'C'] \sqsubseteq \neg D;$

   $['D' + 'C'] \sqsubseteq C; ['NOT' +' D' + 'C'] \sqsubseteq C;$

   $(\neg['D' + 'C'] \sqcap C) \sqsubseteq (\neg['NOT' +' D' + 'C'] \sqcap D);$

   $C$ and $D$ can be complex concepts. Hence, $Rule_{new}: C \sqsubseteq \neg D$ is also included in this rule.

   **A-Box Revision:** $\exists C(a) \in$ ABox $\rightarrow ['D'+'C'](a) \in$ ABox;

B. **Rule 2**: If $Rule_{new}: C \sqsubseteq D$ and $(P \sqsubseteq C \sqcap \neg D) \in$ TBox Then $P \sqsubseteq ['D' + 'C'] \sqcap ['NOT' +' D' + 'C'];$

   $['D' + 'C'] \sqsubseteq D; ['NOT' +' D' + 'C'] \sqsubseteq \neg D;$

   $['D' + 'C'] \sqsubseteq C; ['NOT' +' D' + 'C'] \sqsubseteq C;$

   $(\neg['D' + 'C'] \sqcap C) \sqsubseteq (\neg['NOT' +' D' + 'C'] \sqcap D);$

   **A-Box Revision:** $P(a) \in$ ABox $\rightarrow no\ change;$

C. **Rule 3**: If $Rule_{new}: C(a)$ and $\neg C(a) \in$ ABox Then

   **T-Box Revision:** $C_1 \sqsubseteq C; C_2 \sqsubseteq \neg C;$

   **A-Box Revision:** $C(a) \rightarrow (C_1 \sqcup C_2)(a);$

   $C$ can be complex concepts. Hence, $Rule_{new}: \neg C(a)$ is also included in this rule.

D. **Rule 4**: If $Rule_{new}: C(a)$ and $[(P \sqsubseteq D \sqcap \neg C) \in$ TBox, $P(a) \in$ ABox] Then

   **T-Box Revision:** $C_1 \sqsubseteq C; C_2 \sqsubseteq \neg C; P \sqsubseteq D \sqcap \neg C_1;$

   **A-Box Revision:** $C(a) \rightarrow (C_1 \sqcup C_2)(a);$

   $C$ and $D$ can be complex concepts. Hence, $Rule_{new}: \neg C(a)$ is also included in this rule.

II. **Monotonic Revision Rules**

A. Rule 1: If $Rule_{new}: C \sqsubseteq D_j$ and $(C \sqsubseteq D_{i \neq j}) \in$ TBox Then

   **T-Box Revision**: Add $C \sqsubseteq D_j$. Results $C \sqsubseteq D_j \sqcap D_i$

   **A-Box Revision**: No change.

   $C$ and $D$ can be complex concepts. Hence, $Rule_{new}: \neg C(a)$ is also included in this rule.

B. Rule 2: If $Rule_{new}: C(a)$ and $C(b \neq a) \in$ ABox Then

   **T-Box Revision**: No change

   **A-Box Revision**: Add $C(a)$.

   $C$ can be complex concepts. Hence, $Rule_{new}: \neg C(a)$ is also included in this rule.

# 5. DLOL<sub>IS-A</sub> ARCHITECTURE

## 5.1 Text Pre-Processing Unit

The Text Pre-processing Unit is the $DLOL_{IS-A}$ component that takes unstructured IS-A documents from the interface and then performs the following sequential processes:

- **Step 1** (**Punctuation Processing**): This step is needed to classify IS-A sentences as simple, complex and compound. Other punctuations that do not change the core semantics of the sentences (except for the attitude, emphasis, etc.) are ignored.

- **Step 2** (**Anaphora Resolution**): In the current version of $DLOL_{IS-A}$ the document content anaphora ambiguity is resolved manually since the aim of this work is to understand the inaccuracy that can be attributed to the proposed Ontology Learning technique (and not inaccuracy due to anaphora resolution tool). We tried the tool BART 2007[2] on our dataset but did not get desired accuracy.

- **Step 3** (**Quantifier and Clausal Token Normalization**): Subject/object quantifier, and clausal token variations that are encountered in document content and that are also enlisted in $DLOL_{IS-A}$ variation lists are normalized to their respective common token for unique characterization.

- **Step 4** (**POS Tagging**): After we get a set of normalized sentences from step 3 we then feed them into the Stanford POS Tagger v3.1.3[3] as a first step of characterization of the English IS-A sentences.

- **Step 5** (**IS-A Characterization**): The next step is to characterize the IS-A sentences based on the results of the POS tagger as described in section 4.2. $DLOL_{IS-A}$ then identifies the type of the sentences as simple, complex, or compound (both IS-A type and IS-NOT type).

- **Step 6 (Simple IS-A Normalization)**: The final pre-processing step is to split complex and compound sentences into simple sentences using rules discussed in section 4.2.2 and 4.2.3.

## 5.2 Iterative Ontology Learning Unit

This unit is the second phase of the $DLOL_{IS-A}$ tool and is the core one. In this phase the individual normalized simple IS-A sentences are sequentially mapped to their corresponding DL rules to generate concept definitions. $DLOL_{IS-A}$ uses the Jena API v2.7.3[4] to generate DL definitions as an OWL file. During this process sentences are scanned sequentially as they appear in each document from the beginning of the document till the end. While rules are mapped the corresponding revision rules are also applied to maintain consistency (as discussed in section 4.3). Concept labeling is also done at the same time (as shown within the T-Box and A-Box rules discussed in section 4.2).

## 5.3 Consistent Taxonomy Generation Unit

The Iterative Ontology Learning Unit generates a regular T-Box and A-Box. However, such an ontology is definitional. Hence, the regular ontology needs to be converted into its corresponding generalized form (i.e. axiomatic ontology in the form of concept taxonomy). For that the Consistent Taxonomy Generation Unit calls the Pellet v 1.5.2 reasoner[5] for DL subsumption based taxonomy generation. It also validates the consistency of the generated taxonomy at the same time.

**Table 1. LA Evaluation ($DLOL_{IS-A}$ Vs. Text2Onto)**

| Measure | $DLOL_{IS-A}$ | Text2Onto | $DLOL_{IS-A}$ Gain |
|---|---|---|---|
| $LR_A$ | 0.8104 | 0.4265 | 47.37% |
| $LR_B$ | 0.8692 | 0.4266 | 50.92% |
| $LR_C$ | 0.8954 | 0.4304 | 51.93% |
| **$LR_{Mean}$** | **0.8583** | **0.4278** | **50.07%** |
| **$LR_{SD}$** | **0.0355** | **0.0018** | **NA** |
| $LP_A$ | 0.8671 | 0.7261 | 16.26 |
| $LP_B$ | 0.8866 | 0.7619 | 14.06% |
| $LP_C$ | 0.9072 | 0.7738 | 14.70% |
| **$LP_{Mean}$** | **0.8870** | **0.7539** | **15.01%** |
| **$LP_{SD}$** | **0.0163** | **0.0201** | **NA** |
| $LF_{1A}$ | 0.8377 | 0.5373 | 35.86% |
| $LF_{1B}$ | 0.8777 | 0.5470 | 37.67% |
| $LF_{1C}$ | 0.9012 | 0.5530 | 38.63% |
| **$LF_{1Mean}$** | **0.8722** | **0.5457** | **37.38%** |
| **$LF_{1SD}$** | **0.0262** | **0.0064** | **NA** |

## 6. EVALUATION

The goal of evaluation of the proposed $DLOL_{IS-A}$ tool was twofold: (i) to test the Ontology Learning accuracy of the tool and (ii) to test the coverage of the proposed IS-A characterization.

**A. Accuracy Evaluation**: For the first goal we adopted a Gold Standard evaluation technique widely followed in the community [8, 22]. For conducting our experiments we chose the Wikipedia document on *Mammal* since it is rich in IS-A sentences[6]. We then employed two non ontology expert to create a dataset of IS-A type sentences from these documents and validated the dataset by a linguist. The resulting dataset has 109 IS-A sentences. This dataset was given to three independent ontology engineers that resulted in three different versions (version A, B, and C) of engineered ontology of the same dataset. We adopted two different accuracy metrics: (i) Lexical Accuracy (LA) and (ii) Taxonomic Accuracy (TA).

For LA performance we tested $DLOL_{IS-A}$ in terms of: (i) Lexical Recall (LR), (ii) Lexical Precision (LP), and (iii) Lexical F1 measure ($LF_1$). The lexical accuracy provides an understanding of the agreement between $DLOL_{IS-A}$ and the three ontology engineers in terms of choosing a particular term in the dataset as a concept. While LR measures how much the $DLOL_{IS-A}$ generated

---

[2] http://www.bart-coref.org/

[3] http://nlp.stanford.edu/software/tagger.shtml

[4] http://jena.apache.org/

[5] http://clarkparsia.com/pellet/

[6] en.wikipedia.org/wiki/mammal

**Table 2A. TA Evaluation of DLOL$_{IS-A}$**

| Measures | DLOL$_{IS-A}$ |
|---|---|
| TP$_A$/TR$_A$/TF$_A$/TF'$_A$ | 0.8897 / 0.7634 / 0.8218 / 0.8438 |
| TP$_B$/TR$_B$/TF$_B$/TF'$_B$ | 0.9059 / 0.7445 / 0.8173 / 0.8505 |
| TP$_C$/TR$_C$/TF$_C$/TF'$_C$ | 0.8685 / 0.794 / 0.8295 / 0.8666 |
| TA$_{Mean}$ | **0.888 / 0.7673 / 0.8228 / 0.8536** |
| TA$_{SD}$ | **0.0153 / 0.0203 / 0.005 / 0.0095** |

**Table 2B. TA Evaluation of Text2Onto**

| Measures | Text2Onto |
|---|---|
| TP$_A$/TR$_A$/TF$_A$/TF'$_A$ | 1 / 0.2455 / 0.3942 / 0.4129 |
| TP$_B$/TR$_B$/TF$_B$/TF'$_B$ | 1 / 0.1566 / 0.2709 / 0.3334 |
| TP$_C$/TR$_C$/TF$_C$/TF'$_C$ | 0.9696 / 0.2727 / 0.4257 / 0.4313 |
| TA$_{Mean}$ | **0.9898 / 0.2249 / 0.3636 / 0.3925** |
| TA$_{SD}$ | **0.0143 / 0.0495 / 0.0667 / 0.0424** |

**Table 3: Pair-wise Expert Agreement**

| Expert Pair | DLOL$_{IS-A}$ Cosine Similarity ($^\theta$) |
|---|---|
| Expert A Vs. Expert B | 3.438 |
| Expert B Vs. Expert C | 1.146 |
| Expert A Vs. Expert C | 3.341 |

**Table 4: Domain-specific Coverage Recall (CR)**

| Domain | I | II | III | IV | V | VI |
|---|---|---|---|---|---|---|
| #Sentence | 112 | 32 | 27 | 8 | 7 | 14 |
| # Accept | 104 | 32 | 25 | 7 | 6 | 14 |
| # Pure Accept | 102 | 30 | 11 | 3 | 2 | 8 |
| # Mix Accept | 2 | 2 | 14 | 4 | 4 | 6 |
| # Reject | 8 | 0 | 2 | 1 | 1 | 0 |
| # DLOL$_{IS-A}$ Ft | 3 | 0 | 2 | 1 | 1 | 0 |
| # POS Tag. Ft | 5 | 0 | 0 | 0 | 0 | 0 |
| CR$_O$ | 0.928 | 1 | 0.925 | 0.875 | 0.857 | 1 |
| CR$_E$ | 0.973 | 1 | 0.925 | 0.875 | 0.857 | 1 |

ontology has been able to agree with each of the engineered ontologies, on the other hand LP measures how much the DLOL$_{IS-A}$ generated ontology disagrees with each of the engineered ontologies. The LF$_1$ is the Harmonic Mean of LR and LP. We have compared DLOL$_{IS-A}$ with the popularly chosen baseline tool, Text2Onto, w.r.t the LA measures. Text2Onto is a light-weight Ontology Learning tool. The observations are recorded in table 1. DLOL$_{IS-A}$ learned a total of 153 concepts from the Mammal dataset out of which 142 were primitive concepts and 11 were derived concepts in the T-Box. DLOL$_{IS-A}$ achieves an average LR of 0.8583 while the average LP was observed to be 0.8870. In comparison, Text2Onto performs reasonably well in terms of average LP (0.7539) although it did not fare very well in LR (0.4278). This suggests that linguistic analysis based heavy-weight Ontology Learning can significantly improve the agreement between learned ontology and expert engineered ontology on the same dataset (for DLOL$_{IS-A}$ the average LR improvement is ~50% w.r.t. our chosen baseline).

For TA performance we tested DLOL$_{IS-A}$ in terms of: (i) Taxonomic Precision (TP), (ii) Taxonomic Recall (TR), (iii) Taxonomic F-Measure (TF), and (iv) Taxonomic F'-Measure (TF') [8]. TA is a measure to understand the degree of intersection of the DLOL$_{IS-A}$ generated taxonomy and each of the expert engineered taxonomies. In this sense TA is a better measure than LA since it essentially is a special case of structural subsumption reasoning between two ontologies to check whether one of the ontologies satisfies the other. If an ontology O$_1$ completely satisfies a reference ontology O$_2$ then the TP is of O$_1$ is 100%. If O$_2$ also completely satisfies O$_1$ then the TR of O$_1$ is 100% as well. The TA performance observations of DLOL$_{IS-A}$ are recorded in table 2A[7]. Text2Onto was not generating comparable taxonomy on the chosen dataset. The mean TP was high (~99%) since the taxonomy was very small (no. of concepts was 3). However, the mean TR of Text2Onto was found to be ~22% (as compared to DLOL$_{IS-A}$ mean TR of ~76%). We then performed an agreement test between the performance DLOL$_{IS-A}$ w.r.t each of the expert engineered ontologies. We modeled the accuracy of DLOL$_{IS-A}$ as a vector with 7 features (3 of LA and 4 of TA). This resulted in generating 3 such vectors corresponding to the results w.r.t. the 3 experts. We then calculated the cosine similarity between each of these vectors as recorded in table 3. We can observe that DLOL$_{IS-A}$ had very strong agreement with expert B and C (1.146$^0$).

**Characterization Coverage:** For evaluating how much the proposed IS-A characterization can cover IS-A sentences from different domains we created a larger dataset spanning 6 different domains – Animal (I), Tourism (II), Drug (III), Plant (IV), Emotion (V), and Miscellaneous (VI). The dataset consists of purely ISA and mixed IS-A sentences (i.e. at least one clause exists that is non IS-A). The dataset statistics can be seen in table 4[8]. Out of a total of 200 sentences ($N_S$) DLOL$_{IS-A}$ could accept correctly 188 sentences (*#Accept*). However, it rejected 12 sentences (~6%) out of which 5 got rejected due to wrong POS tagging (*# PosTag.Ft*). The rest (7) was rejected because of coverage failure. The effective Coverage Recall is: $R_E = \frac{\#Accept + \# POS\ Tag.Ft}{N_S}$. The mean CR$_E$ was observed to be 0.938.

## 7. CONCLUSION
In this paper we propose a characterization of IS-A sentences that can extensively cover most types of IS-A sentences. We then proposed a heavy-weight Ontology Learning tool called DLOL$_{IS-A}$ that automatically generates an axiomatic ontology on a given IS-A corpus content. The approach does not require any human intervention during concept hierarchy construction and do not depend on any external ontology. We achieve promising results. To the best of our knowledge this is the first attempt on DL based Ontology Learning.

## 8. ACKNOWLEDGMENTS
We would like to thank Mandar Mitra (Indian Statistical Institute) and Adit Nath Sarkar (DA-IICT) for their constant help.

---

[7] TP$_i$: TP w.r.t i-th expert engineer (where i = A,B,C).

[8] Ft: Abbrev. for Fault; CR$_O$ & CR$_E$: Coverage Recall – Overall and Coverage Recall – Effective resp.